\documentclass[11pt,a4paper]{article}
\usepackage[hyperref]{emnlp2018}
\usepackage{times}
\usepackage{latexsym}

\usepackage{graphicx}
\usepackage{wrapfig}
\usepackage{url}
\usepackage{amsmath}
\usepackage{amssymb}
\usepackage{color,xcolor,colortbl}
\usepackage{enumitem}
\usepackage{algorithm}
\usepackage{algorithmic}
\usepackage[font={small}]{caption}
\usepackage{bm,bbm}
\usepackage{booktabs}
\usepackage{mathtools}
\usepackage{array}
\usepackage{multirow}

\DeclareMathOperator*{\argmax}{arg\,max\,}
\newcommand\numberthis{\addtocounter{equation}{1}\tag{\theequation}}

\definecolor{darkblue}{rgb}{0.0, 0.0, 0.55}
\newenvironment{fontppl}{\fontfamily{ppl}\selectfont}{\par} % Palatino

\aclfinalcopy % Uncomment this line for the final submission
\def\aclpaperid{1324} %  Enter the acl Paper ID here

\title{Adapting the Neural Encoder-Decoder Framework from Single to Multi-Document Summarization}

\author{Logan Lebanoff, Kaiqiang Song and Fei Liu\\ 
  Department of Computer Science\\
  University of Central Florida, 
  Orlando, FL 32816, USA\\
  {\tt \{loganlebanoff, kqsong\}@knights.ucf.edu \quad feiliu@cs.ucf.edu}}

\date{}

\begin{document}
\maketitle
\begin{abstract}

Generating a text abstract from a set of documents remains a challenging task.
The neural encoder-decoder framework has recently been exploited to summarize single documents, but its success can in part be attributed to the availability of large parallel data automatically acquired from the Web.
In contrast, parallel data for multi-document summarization are scarce and costly to obtain.
There is a pressing need to adapt an encoder-decoder model trained on single-document summarization data to work with multiple-document input.
In this paper, we present an initial investigation into a novel adaptation method.
It exploits the maximal marginal relevance method to select representative sentences from multi-document input, and leverages an abstractive encoder-decoder model to fuse disparate sentences to an abstractive summary. 
The adaptation method is robust and itself requires no training data. 
Our system compares favorably to state-of-the-art extractive and abstractive approaches judged by automatic metrics and human assessors.

\end{abstract}

\section{Introduction}
\label{sec:intro}
Neural abstractive summarization has primarily focused on summarizing short texts written by single authors.
For example, \emph{sentence summarization} seeks to reduce the first sentence of a news article to a title-like summary~\cite{Rush:2015,Nallapati:2016,Takase:2016,Song:2018}; \emph{single-document summarization} (\textbf{SDS}) focuses on condensing a news article to a handful of bullet points~\cite{Paulus:2017,See:2017}.
These summarization studies are empowered by large parallel datasets automatically harvested from online news outlets, including Gigaword~\cite{Rush:2015}, CNN/Daily Mail~\cite{Hermann:2015}, NYT~\cite{Sandhaus:2008}, and Newsroom~\cite{Grusky:2018}.

To date, \emph{multi-document summarization} (\textbf{MDS}) has not yet fully benefited from the development of neural encoder-decoder models.
MDS seeks to condense a set of documents likely written by multiple authors to a short and informative summary. 
It has practical applications, such as summarizing product reviews~\cite{Gerani:2014}, student responses to post-class questionnaires~\cite{Luo:2015,Luo:2016:NAACL}, and sets of news articles discussing certain topics~\cite{Hong:2014}. 
State-of-the-art MDS systems are mostly extractive~\cite{Nenkova:2011}.
Despite their promising results, such systems cannot perform text abstraction, e.g., paraphrasing, generalization, and sentence fusion~\cite{Jing:1999}.
Further, annotated MDS datasets are often scarce, containing only hundreds of training pairs (see Table~\ref{tab:example}).
The cost to create ground-truth summaries from multiple-document inputs can be prohibitive.
The MDS datasets are thus too small to be used to train neural encoder-decoder models with millions of parameters without overfitting.

\begin{table}
\setlength{\tabcolsep}{4pt}
\renewcommand{\arraystretch}{1.2}
\centering
\begin{scriptsize}
\begin{fontppl}
\begin{tabular}{|l|l|l|c|}
\hline
\textsc{\textbf{Dataset}} & \textsc{\textbf{Source}} & \textsc{\textbf{Summary}} & \#\textsc{\textbf{Pairs}}\\
\hline
\hline
\textbf{Gigaword} & \textbf{the first sentence}  & \textbf{8.3 words} & \multirow{2}{*}{\textbf{4 Million}} \\
\cite{Rush:2015} & of a news article & title-like & \\
\hline
\textbf{CNN/Daily Mail} & \multirow{2}{*}{\textbf{a news article}} & \textbf{56 words}  & \multirow{2}{*}{\textbf{312 K}} \\
\cite{Hermann:2015} &  & multi-sent & \\
\hline
\textbf{TAC (08-11)} & \textbf{10 news articles}  & \textbf{100 words} & \multirow{2}{*}{\textbf{728}}\\
(Dang et al., 2008)\nocite{Dang:2008} & related to a topic & multi-sent & \\
\hline
\textbf{DUC (03-04)} & \textbf{10 news articles}  & \textbf{100 words} & \multirow{2}{*}{\textbf{320}}\\
\cite{Over:2004} & related to a topic & multi-sent & \\
\hline
\end{tabular}
\end{fontppl}
\end{scriptsize}
\caption{A comparison of datasets available for sent. summarization (Gigaword), single-doc (CNN/DM) and multi-doc summarization (DUC/TAC). The labelled data for multi-doc summarization are much less.}
\label{tab:example}
\vspace{-0.2in}
\end{table}
% \rotatebox[origin=c]{90}{}

A promising route to generating an abstractive summary from a multi-document input is to apply a neural encoder-decoder model trained for single-document summarization to a ``\emph{mega-document}'' created by concatenating all documents in the set at test time.
Nonetheless, such a model may not scale well for two reasons.
First, identifying important text pieces from a mega-document can be challenging for the encoder-decoder model, which is trained on single-document summarization data where the summary-worthy content is often contained in the first few sentences of an article. 
This is not the case for a mega-document.
Second, redundant text pieces in a mega-document can be repeatedly used for summary generation under the current framework.
The attention mechanism of an encoder-decoder model~\cite{Bahdanau:2014} is position-based and lacks an awareness of semantics.
If a text piece has been attended to during summary generation, it is unlikely to be used again.
However, the attention value assigned to a similar text piece in a different position is not affected.
The same content can thus be repeatedly used for summary generation.
These issues may be alleviated by improving the encoder-decoder architecture and its attention mechanism~\cite{Cheng:2016,Tan:2017}. 
However, in these cases the model has to be re-trained on large-scale MDS datasets that are not available at the current stage.
There is thus an increasing need for a lightweight adaptation of an encoder-decoder model trained on SDS datasets to work with multi-document inputs at test time.

In this paper, we present a novel adaptation method, named PG-MMR, to generate abstracts from multi-document inputs.
The method is robust and requires no MDS training data. 
% itself does not require any additional training data outside of the abundant SDS data.
It combines a recent neural encoder-decoder model (PG for Pointer-Generator networks; See et al., 2017\nocite{See:2017}) that generates abstractive summaries from single-document inputs with a strong extractive summarization algorithm (MMR for Maximal Marginal Relevance; Carbonell and Goldstein, 1998\nocite{Carbonell:1998}) that identifies important source sentences from multi-document inputs.
The PG-MMR algorithm iteratively performs the following.
It identifies a handful of the most important sentences from the mega-document.
The attention weights of the PG model are directly modified to focus on these important sentences when generating a summary sentence.
Next, the system re-identifies a number of important sentences, but the likelihood of choosing certain sentences is reduced based on their similarity to the partially-generated summary, thereby reducing redundancy.
Our research contributions include the following:
\begin{itemize}[topsep=3pt,itemsep=-1pt,leftmargin=*]

\item we present an investigation into a novel adaptation method of the encoder-decoder framework from single- to multi-document summarization.
To the best of our knowledge, this is the first attempt to couple the maximal marginal relevance algorithm with pointer-generator networks for multi-document summarization;

\item we demonstrate the effectiveness of the proposed method through extensive experiments on standard MDS datasets. 
Our system compares favorably to state-of-the-art extractive and abstractive summarization systems measured by both automatic metrics and human judgments.

\end{itemize}

\section{Related Work}
\label{sec:related_work}

Popular methods for multi-document summarization have been extractive. 
Important sentences are extracted from a set of source documents and optionally compressed to form a summary~\cite{Daume:2002,Zajic:2007,Gillick:2009:NAACL,Galanis:2010,Kirkpatrick:2011,Li:2013:EMNLP,Thadani:2013,Wang:2013,Yogatama:2015:EMNLP,Filippova:2015,Durrett:2016}.
In recent years neural networks have been exploited to learn word/sentence representations for single- and multi-document summarization~\cite{Cheng:2016,Cao:2017,Isonuma:2017,Yasunaga:2017,Narayan:2018}. 
These approaches remain extractive; and despite encouraging results, summarizing a large quantity of texts still requires sophisticated abstraction capabilities such as generalization, paraphrasing and sentence fusion.

Prior to deep learning, abstractive summarization has been investigated~\cite{Barzilay:1999,Carenini:2008,Ganesan:2010,Gerani:2014,Fabbrizio:2014,Pighin:2014,Bing:2015,Liu:2015:NAACL,Liao:2018}. 
These approaches construct domain templates using a text planner or an open-IE system and employ a natural language generator for surface realization.
Limited by the availability of labelled data, experiments are often performed on small domain-specific datasets.

Neural abstractive summarization utilizing the encoder-decoder architecture has shown promising results but studies focus primarily on single-document summarization~\cite{Nallapati:2016,Kikuchi:2016,Chen:2016,Miao:2016,Tan:2017,Zeng:2017,Zhou:2017,Paulus:2017,See:2017,Gehrmann:2018}.
The pointing mechanism~\cite{Gulcehre:2016,Gu:2016} allows a summarization system to both copy words from the source text and generate new words from the vocabulary.
Reinforcement learning is exploited to directly optimize evaluation metrics~\cite{Paulus:2017,Kryscinski:2018,Chen:2018:ACL}.
These studies focus on summarizing single documents in part because the training data are abundant. 

The work of Baumel et al.~\shortcite{Baumel:2018} and Zhang et al.~\shortcite{Zhang:2018} are related to ours. 
In particular, Baumel et al.~\shortcite{Baumel:2018} propose to extend an abstractive summarization system to generate query-focused summaries;
Zhang et al.~\shortcite{Zhang:2018} add a document set encoder to their hierarchical summarization framework.
With these few exceptions, little research has been dedicated to investigate the feasibility of extending the encoder-decoder framework to generate abstractive summaries from multi-document inputs, where available training data are scarce.

This paper presents some first steps towards the goal of extending the encoder-decoder model to a multi-document setting. 
We introduce an adaptation method combining the pointer-generator (PG) networks~\cite{See:2017} and the maximal marginal relevance (MMR) algorithm~\cite{Carbonell:1998}.
The PG model, trained on SDS data and detailed in Section \S\ref{sec:PG_networks}, is capable of generating document abstracts by performing text abstraction and sentence fusion. However, if the model is applied at test time to summarize multi-document inputs, there will be limitations. 
Our PG-MMR algorithm, presented in Section \S\ref{sec:our_approach}, teaches the PG model to effectively recognize important content from the input documents, hence improving the quality of abstractive summaries, all without requiring any training on multi-document inputs.

\begin{figure*}
\centering
\includegraphics[width=5.8in]{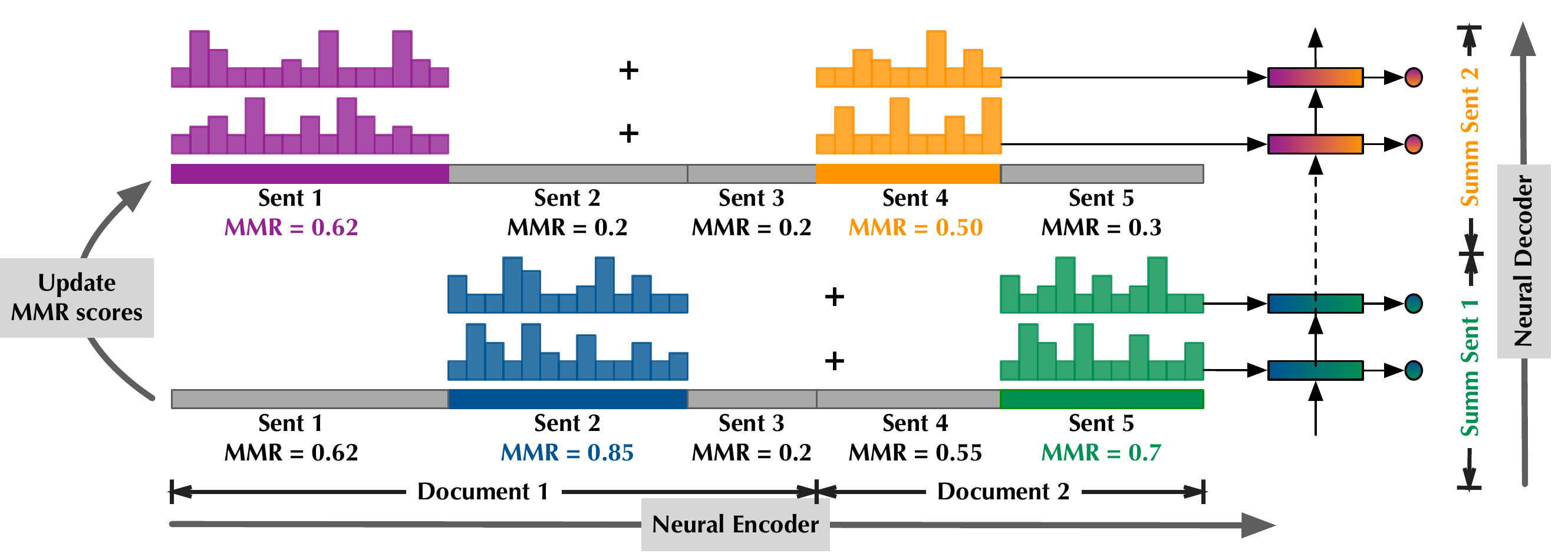}
\caption{System framework. The PG-MMR system uses K highest-scored source sentences (in this case, K=2) to guide the PG model to generate a summary sentence. All other source sentences are ``muted'' in this process. Best viewed in color. }
\label{fig:framework}
\vspace{-0.1in}
\end{figure*}

\section{Limits of the Encoder-Decoder Model}
\label{sec:PG_networks}

The encoder-decoder architecture has become the \emph{de facto} standard for neural abstractive summarization~\cite{Rush:2015}.
The encoder is often a bidirectional LSTM~\cite{Hochreiter:1997} converting the input text to a set of hidden states $\{\mathbf{h}_i^e\}$, one for each input word, indexed by $i$.
The decoder is a unidirectional LSTM that generates a summary by predicting one word at a time.
The decoder hidden states are represented by $\{\mathbf{h}_t^d\}$, indexed by $t$.
For sentence and single-document summarization~\cite{Nallapati:2016,Paulus:2017,See:2017}, the input text is treated as a sequence of words, and the model is expected to capture the source syntax inherently.
\begin{align*}
& {e}_{t,i} = \mathbf{v}^\top \tanh(\mathbf{W}^e [\mathbf{h}_t^d || \mathbf{h}_i^e || \widetilde{\alpha}_{t,i}] + \mathbf{b}^e)
\numberthis\label{equ:e_t_i}\\
& \alpha_{t,i} = \mbox{softmax}(e_{t,i})
\numberthis\label{equ:alpha_t_i}\\
& \widetilde{\alpha}_{t,i} = \textstyle\sum_{t'=0}^{t-1} \alpha_{t',i}
\numberthis\label{equ:alpha_tilde}
\end{align*}

The attention weight $\alpha_{t,i}$ measures how important the $i$-th input word is to generating the $t$-th output word (Eq.~(\ref{equ:e_t_i}-\ref{equ:alpha_t_i})).
Following~\cite{See:2017}, $\alpha_{t,i}$ is calculated by measuring the strength of interaction between the decoder hidden state $\mathbf{h}_t^d$, the encoder hidden state $\mathbf{h}_i^e$, and the \emph{cumulative} attention $\widetilde{\alpha}_{t,i}$ (Eq.~(\ref{equ:alpha_tilde})).
$\widetilde{\alpha}_{t,i}$ denotes the {cumulative} attention that the $i$-th input word receives up to time step $t$-1.
A large value of $\widetilde{\alpha}_{t,i}$ indicates the $i$-th input word has been used prior to time $t$ and it is unlikely to be used again for generating the $t$-th output word.

A context vector ($\mathbf{c}_t$) is constructed (Eq.~(\ref{equ:c_t})) to summarize the semantic meaning of the input; it is a weighted sum of the encoder hidden states.
The context vector and the decoder hidden state ($[\mathbf{h}_t^d || \mathbf{c}_t]$) are then used to compute the vocabulary probability $P_{vcb}(w)$ measuring the likelihood of a vocabulary word $w$ being selected as the $t$-th output word (Eq.~(\ref{equ:p_vocab})).\footnote{Here $[\cdot||\cdot]$ represents the concatenation of two vectors. The pointer-generator networks~\cite{See:2017} use two linear layers to produce the vocabulary distribution $P_{vcb}(w)$. We use $\mathbf{W}^y$ and $\mathbf{b}^y$ to denote parameters of both layers.}
\begin{align*}
& \mathbf{c}_t = \textstyle\sum_{i} \alpha_{t,i} \mathbf{h}_i^e
\numberthis\label{equ:c_t}\\
& P_{vcb}(w) = \text{softmax}(\mathbf{W}^y[\mathbf{h}_t^d || \mathbf{c}_t] + \mathbf{b}^y)
\numberthis\label{equ:p_vocab}
\end{align*}

In many encoder-decoder models, a ``switch'' is estimated ($p_{gen} \in$ [0,1]) to indicate whether the system has chosen to select a word from the vocabulary or to copy a word from the input text (Eq.~(\ref{equ:p_gen})).
The switch is computed using a feedforward layer with $\sigma$ activation over $[\mathbf{h}_t^d || \mathbf{c}_t || \mathbf{y}_{t-1}]$, where $\mathbf{y}_{t-1}$ is the embedding of the output word at time $t$-1.
The attention weights ($\alpha_{t,i}$) are used to compute the copy probability (Eq.~(\ref{equ:p_w})).
If a word $w$ appears once or more in the input text, its copy probability ($\sum_{i:w_i = w} \alpha_{t,i}$) is the sum of the attention weights over all its occurrences. 
The final probability $P(w)$ is a weighted combination of the vocabulary probability and the copy probability. 
A cross-entropy loss function can often be used to train the model end-to-end.
{\medmuskip=1mu
\thinmuskip=1mu
\thickmuskip=1mu
\nulldelimiterspace=0pt
\scriptspace=0pt
\begin{align*}
& p_{gen} = \sigma(\mathbf{w}^z[\mathbf{h}_t^d || \mathbf{c}_t || \mathbf{y}_{t-1}]) + b^z)
\numberthis\label{equ:p_gen}\\
& P(w) = p_{gen} P_{vcb}(w) + (1 - p_{gen}) \sum_{i:w_i = w} \alpha_{t,i}
\numberthis\label{equ:p_w}
\end{align*}}

\vspace{-0.1in}
To thoroughly understand the aforementioned encoder-decoder model, we divide its model parameters into four groups.
They include
\begin{itemize}[topsep=3pt,itemsep=-1pt,leftmargin=*]
\item parameters of the encoder and the decoder;
\item $\{\mathbf{w}^z, b^z\}$ for calculating the ``switch'' (Eq.~(\ref{equ:p_gen}));
\item $\{\mathbf{W}^y, \mathbf{b}^y\}$ for calculating $P_{vcb}(w)$ (Eq.~(\ref{equ:p_vocab}));
\item $\{\mathbf{v}, \mathbf{W}^e, \mathbf{b}^e\}$ for attention weights (Eq.~(\ref{equ:e_t_i})).
\end{itemize}
By training the encoder-decoder model on single-document summarization (SDS) data containing a large collection of news articles paired with summaries~\cite{Hermann:2015}, these model parameters can be effectively learned.

However, at test time, we wish for the model to generate abstractive summaries from \emph{multi-document inputs}.
This brings up two issues. 
First, the parameters are ineffective at identifying salient content from multi-document inputs. 
Humans are very good at identifying representative sentences from a set of documents and fusing them into an abstract.
However, this capability is not supported by the encoder-decoder model. 
Second, the attention mechanism is based on input word positions but not their semantics. 
It can lead to redundant content in the multi-document input being repeatedly used for summary generation.
We conjecture that both aspects can be addressed by introducing an ``external'' model that selects representative sentences from multi-document inputs and dynamically adjusts the sentence importance to reduce summary redundancy. 
This external model is integrated with the encoder-decoder model to generate abstractive summaries using selected representative sentences.
In the following section we present our adaptation method for multi-document summarization.

\section{Our Method}
\label{sec:our_approach}

\noindent\textbf{Maximal marginal relevance.}
Our adaptation method incorporates the maximal marginal relevance algorithm (MMR; Carbonell and Goldstein, 1998\nocite{Carbonell:1998}) into pointer-generator networks (PG; See et al., 2017\nocite{See:2017}) by adjusting the network's attention values.
MMR is one of the most successful extractive approaches and, despite its straightforwardness, performs on-par with state-of-the-art systems~\cite{Luo:2015,Yogatama:2015:EMNLP}.
At each iteration, MMR selects one sentence from the document ($D$) and includes it in the summary ($S$) until a length threshold is reached.
The selected sentence ($s_i$) is the most important one amongst the remaining sentences and it has the least content overlap with the current summary.
In the equation below, $\mbox{Sim}_1(s_i,D)$ measures the similarity of the sentence $s_i$ to the document. It serves as a proxy of sentence importance, since important sentences usually show similarity to the centroid of the document.
$\max_{s_j \in S} \mbox{Sim}_2(s_i, s_j)$ measures the maximum similarity of the sentence $s_i$ to each of the summary sentences, acting as a proxy of redundancy. 
$\lambda$ is a balancing factor.
{\medmuskip=1mu
\thinmuskip=1mu
\thickmuskip=1mu
\nulldelimiterspace=0pt
\scriptspace=0pt
\begin{align*}
&\argmax_{s_i \in D \setminus S} \big[\underbrace{\lambda \mbox{Sim}_1(s_i,D)\vphantom{\max_{s_j \in S} \mbox{Sim}_2}}_{\mbox{importance}} - \underbrace{(1-\lambda) \max_{s_j \in S} \mbox{Sim}_2(s_i, s_j)}_{\mbox{redundancy}} \big]
% \numberthis\label{equ:mmr}
\end{align*}}

\vspace{-0.05in}
Our PG-MMR describes an iterative framework for summarizing a multi-document input to a summary consisting of multiple sentences.
At each iteration, PG-MMR follows the MMR principle to select the K highest-scored source sentences; they serve as the basis for PG to generate a summary sentence. 
After that, the scores of all source sentences are updated based on their importance and redundancy.
Sentences that are highly similar to the partial summary receive lower scores.
Selecting K sentences via the MMR algorithm helps the PG system to effectively identify salient source content that has not been included in the summary.

\vspace{0.05in}
\noindent\textbf{Muting.}
To allow the PG system to effectively utilize the K source sentences without retraining the neural model, we dynamically adjust the PG attention weights ($\alpha_{t,i}$) at test time. 
Let $\mbox{S}_k$ represent a selected sentence.
The attention weights of the words belonging to \{S$_k$\}$_{k=1}^K$ are calculated as before (Eq.~(\ref{equ:alpha_t_i})).
However, words in other sentences are forced to receive zero attention weights ($\alpha_{t,i}$=0), and all $\alpha_{t,i}$ are renormalized (Eq.~(\ref{equ:alpha_t_i_new})). 
{\medmuskip=1mu
\thinmuskip=1mu
\thickmuskip=1mu
\nulldelimiterspace=0pt
\scriptspace=0pt
\begin{align*}
\alpha_{t,i}^{\mbox{\scriptsize new}} = 
\begin{cases}
\alpha_{t,i} & i \in \{\mbox{S}_k\}_{k=1}^K \\
0 & \mbox{otherwise} \numberthis\label{equ:alpha_t_i_new}
\end{cases}
\end{align*}}
It means that the remaining sentences are ``muted'' in this process. 
In this variant, the sentence importance does not affect the original attention weights, other than muting. 

In an alternative setting, the sentence salience is multiplied with the word salience and renormalized (Eq.~(\ref{equ:alpha_t_i_new_alternative})). 
PG uses the reweighted alpha values to predict the next summary word.
{\medmuskip=1mu
\thinmuskip=1mu
\thickmuskip=1mu
\nulldelimiterspace=0pt
\scriptspace=0pt
\begin{align*}
\alpha_{t,i}^{\mbox{\scriptsize new}} = 
\begin{cases}
\alpha_{t,i} \mbox{MMR}(\mbox{S}_k) & i \in \{\mbox{S}_k\}_{k=1}^K \\
0 & \mbox{otherwise} \numberthis\label{equ:alpha_t_i_new_alternative}
\end{cases}
\end{align*}}

\vspace{-0.15in}
\noindent\textbf{Sentence Importance.}
To estimate sentence importance $\mbox{Sim}_1(s_i,D)$, we introduce a supervised regression model in this work.
Importantly, the model is trained on single-document summarization datasets where training data are abundant. 
At test time, the model can be applied to identify important sentences from multi-document input. 
Our model determines sentence importance based on four indicators, inspired by how humans identify important sentences from a document set. 
They include (a) sentence length, (b) its absolute and relative position in the document, (c) sentence quality, and (d) how close the sentence is to the main topic of the document set. 
These features are considered to be important indicators in previous extractive summarization framework~\cite{Galanis:2010,Hong:2014}. 

Regarding the sentence quality (c), we leverage the PG model to build the sentence representation.
We use the bidirectional LSTM encoder to encode any source sentence to a vector representation.
$[\overrightarrow{\mathbf{h}_N^e} || \overleftarrow{\mathbf{h}_1^e}]$ is the concatenation of the last hidden states of the forward and backward passes.
A document vector is the average of all sentence vectors. 
We use the document vector and the cosine similarity between the document and sentence vectors as indicator (d).
A support vector regression model is trained on (sentence, score) pairs where the training data are obtained from the CNN/Daily Mail dataset.
The target importance score is the ROUGE-L recall of the sentence compared to the ground-truth summary.
Our model architecture leverages neural representations of sentences and documents, they are data-driven and not restricted to a particular domain.
% not specifically designed for a particular domain.

\vspace{0.05in}
\noindent\textbf{Sentence Redundancy.}
To calculate the redundancy of the sentence ($\max_{s_j \in S} \mbox{Sim}_2(s_i, s_j)$), we compute the ROUGE-L precision, which measures the longest common subsequence between a source sentence and the partial summary (consisting of all sentences generated thus far by the PG model), divided by the length of the source sentence.
A source sentence yielding a high ROUGE-L precision is deemed to have significant content overlap with the partial summary. 
It will receive a low MMR score and hence is less likely to serve as basis for generating future summary sentences.

Alg.~\ref{alg:pg_mmr} provides an overview the PG-MMR algorithm and Fig.~\ref{fig:framework} is a graphical illustration.
The MMR scores of source sentences are updated after each summary sentence is generated by the PG model.
Next, a different set of highest-scored sentences are used to guide the PG model to generate the next summary sentence.
``Muting'' the remaining source sentences is important because it helps the PG model to focus its attention on the most significant source content.
The code for our model is publicly available to further MDS research.\footnote{\href{https://github.com/ucfnlp/multidoc_summarization}{https://github.com/ucfnlp/multidoc\_summarization}}
\setlength{\textfloatsep}{0pt}
\begin{algorithm}[t]
\begin{algorithmic}[1]
\REQUIRE SDS data; MDS source sentences \{S$_i$\}
\STATE Train the PG model on SDS data
\STATE \COMMENT{\textcolor{darkblue}{$\mathcal{I}$(S$_i$) and $\mathcal{R}$(S$_i$) are the importance and redundancy scores of the source sentence S$_i$}}
\STATE $\mathcal{I}$(S$_i$) $\leftarrow$ SVR(S$_i$) for all source sentences
\STATE MMR(S$_i$) $\leftarrow \lambda\mathcal{I}$(S$_i$) for all source sentences
\STATE Summary $\leftarrow \{\}$ 
% $\quad$ \COMMENT{\textcolor{darkblue}{System summary.}}
\STATE $t \leftarrow$ index of summary words
\WHILE{$t < L_{\mbox{\scriptsize max}}$}
\STATE Find \{S$_k$\}$_{k=1}^K$ with highest MMR scores
\STATE Compute $\alpha_{t,i}^{\mbox{\scriptsize new}}$ based on \{S$_k$\}$_{k=1}^K$ (Eq.~(\ref{equ:alpha_t_i_new}))
\STATE Run PG decoder for one step to get $\{w_t\}$
\STATE Summary $\leftarrow$ Summary + $\{w_t\}$
\IF{$w_t$ is the period symbol}
\STATE $\mathcal{R}$(S$_i$) $\leftarrow$ Sim(S$_i$, Summary), $\forall i$
\STATE MMR(S$_i$) $\leftarrow \lambda\mathcal{I}$(S$_i$) $- (1-\lambda)\mathcal{R}$(S$_i$), $\forall i$
\ENDIF 
\ENDWHILE
\end{algorithmic}
\caption{The PG-MMR algorithm for summarizing multi-document inputs.}
\label{alg:pg_mmr}
\end{algorithm}
\setlength{\textfloatsep}{15pt}

\section{Experimental Setup}
\label{sec:experiments}

\noindent\textbf{Datasets.} 
We investigate the effectiveness of the PG-MMR method by testing it on standard multi-document summarization datasets~\cite{Over:2004,Dang:2008}.
These include DUC-03, DUC-04, TAC-08, TAC-10, and TAC-11, containing 30/50/48/46/44 topics respectively.
The summarization system is tasked with generating a concise, fluent summary of 100 words or less from a set of 10 documents discussing a topic. 
All documents in a set are chronologically ordered and concatenated to form a mega-document serving as input to the PG-MMR system. 
Sentences that start with a quotation mark or do not end with a period are excluded~\cite{Wong:2008}. 
Each system summary is compared against 4 human abstracts created by NIST assessors.
Following convention, we report results on DUC-04 and TAC-11 datasets, which are standard test sets; DUC-03 and TAC-08/10 are used as a validation set for hyperparameter tuning.\footnote{The hyperparameters for all PG-MMR variants are $K$=7 and $\lambda$=0.6; except for ``w/ BestSummRec'' where $K$=2.}

The PG model is trained for single-document summarization using the CNN/Daily Mail~\cite{Hermann:2015} dataset, containing single news articles paired with summaries (human-written article highlights). 
The training set contains 287,226 articles.
An article contains 781 tokens on average; and a summary contains 56 tokens (3.75 sentences).
During training we use the hyperparameters provided by See et al.~\shortcite{See:2017}.
At test time, the maximum/minimum decoding steps are set to 120/100 words respectively, corresponding to the max/min lengths of the PG-MMR summaries. 
Because the focus of this work is on multi-document summarization (MDS), we do not report results for the CNN/Daily Mail dataset. 

\vspace{0.05in}
\noindent\textbf{Baselines.} 
We compare PG-MMR against a broad spectrum of baselines, including state-of-the-art extractive (`\textit{ext-}') and abstractive (`\textit{abs-}') systems. They are described below.\footnote{We are grateful to Hong et al.~\shortcite{Hong:2014} for providing the summaries generated by Centroid, ICSISumm, DPP systems. These are only available for the DUC-04 dataset.}

\begin{itemize}[topsep=5pt,itemsep=0pt,leftmargin=*]
\begin{footnotesize}
\item \textit{ext-}\textbf{SumBasic}~\cite{Vanderwende:2007} is an extractive approach assuming words occurring frequently in a document set are more likely to be included in the summary;

\item \textit{ext-}\textbf{KL-Sum}~\cite{Haghighi:2009} greedily adds source sentences to the summary if it leads to a decrease in KL divergence;

\item \textit{ext-}\textbf{LexRank}~\cite{Erkan:2004} uses a graph-based approach to compute sentence importance based on eigenvector centrality in a graph representation;
\item \textit{ext-}\textbf{Centroid}~\cite{Hong:2014} computes the importance of each source sentence based on its cosine similarity with the document centroid;
\item \textit{ext-}\textbf{ICSISumm}~\cite{Gillick:2009:UTD} leverages the ILP framework to identify a globally-optimal set of sentences covering the most important concepts in the document set;

\item \textit{ext-}\textbf{DPP}~\cite{Kulesza:2012} selects an optimal set of sentences per the determinantal point processes that balance the coverage of important information and the sentence diversity;

\item \textit{abs-}\textbf{Opinosis}~\cite{Ganesan:2010} generates abstractive summaries by searching for salient paths on a word co-occurrence graph created from source documents;

\item \textit{abs-}\textbf{Extract+Rewrite}~\cite{Song:2018} is a recent approach that scores sentences using LexRank and generates a title-like summary for each sentence using an encoder-decoder model trained on Gigaword data.  

\item \textit{abs-}\textbf{PG-Original}~\cite{See:2017} introduces an encoder-decoder model that encourages the system to copy words from the source text via pointing, while retaining the ability to produce novel words through the generator. 

\end{footnotesize}
\end{itemize}

\begin{table}[t]
\setlength{\tabcolsep}{5pt}
\renewcommand{\arraystretch}{1.1}
\centering
\begin{small}
\begin{tabular}{|l|rrr|}
\hline
& \multicolumn{3}{c|}{\textbf{DUC-04}}\\
\textbf{System} & \textbf{R-1} & \,\,\,\textbf{R-2} & \textbf{R-SU4} \\
\hline
\hline
SumBasic{\scriptsize~\cite{Vanderwende:2007}} & 29.48 & 4.25 & 8.64\\
KLSumm{\scriptsize~(Haghighi et al., 2009)\nocite{Haghighi:2009}} & 31.04 & 6.03 & 10.23 \\
LexRank{\scriptsize~\cite{Erkan:2004}} & 34.44 & 7.11 & 11.19 \\
Centroid{\scriptsize~\cite{Hong:2014}} & 35.49 & 7.80 & 12.02 \\
ICSISumm{\scriptsize~\cite{Gillick:2009:NAACL}} & 37.31 & 9.36 & 13.12 \\
DPP{\scriptsize~\cite{Kulesza:2012}} & 38.78 & 9.47 & 13.36 \\
Extract+Rewrite{\scriptsize~\cite{Song:2018}} & 28.90 & 5.33 & 8.76 \\
Opinosis{\scriptsize~\cite{Ganesan:2010}} & 27.07 & 5.03 & 8.63 \\
PG-Original{\scriptsize~\cite{See:2017}} & 31.43 & 6.03 & 10.01\\
\hline
PG-MMR w/ SummRec & 34.57 & 7.46 & 11.36\\
PG-MMR w/ SentAttn & 36.52 & 8.52 & 12.57\\
PG-MMR w/ Cosine (\emph{default}) & 36.88 & \textbf{8.73} & 12.64\\
\hline
PG-MMR w/ BestSummRec & 36.42 & \textbf{9.36} & 13.23\\
\hline
\end{tabular}
\end{small}
\caption{ROUGE results on the DUC-04 dataset. 
}
\label{tab:results_duc04}
\vspace{-0.1in}
\end{table}

\begin{table}[t]
\setlength{\tabcolsep}{4pt}
\renewcommand{\arraystretch}{1.1}
\centering
\begin{small}
\begin{tabular}{|l|rrr|}
\hline
& \multicolumn{3}{c|}{\textbf{TAC-11}}\\
\textbf{System} & \textbf{R-1} & \textbf{R-2} & \textbf{R-SU4} \\
\hline
\hline
SumBasic{\scriptsize~\cite{Vanderwende:2007}} & 31.58 & 6.06 & 10.06\\
KLSumm{\scriptsize~(Haghighi et al., 2009)\nocite{Haghighi:2009}} & 31.23 & 7.07 & 10.56 \\
LexRank{\scriptsize~\cite{Erkan:2004}} & 33.10 & 7.50 & 11.13 \\
Extract+Rewrite{\scriptsize~\cite{Song:2018}} & 29.07 & 6.11 & 9.20\\
Opinosis{\scriptsize~\cite{Ganesan:2010}} & 25.15 & 5.12 & 8.12\\
PG-Original{\scriptsize~\cite{See:2017}} & 31.44 & 6.40 & 10.20\\
\hline
PG-MMR w/ SummRec & 35.06 & 8.72 & 12.39\\
PG-MMR w/ SentAttn & 37.01 & 10.43 & 13.85\\
PG-MMR w/ Cosine (\emph{default}) & 37.17 & \textbf{10.92} & 14.04\\
\hline
PG-MMR w/ BestSummRec & 40.44 & \textbf{14.93} & 17.61\\
\hline
\end{tabular}
\end{small}
\caption{ROUGE results on the TAC-11 dataset. 
}
\label{tab:results_tac11}
\vspace{-0.1in}
\end{table}

\section{Results}
\label{sec:results}

Having described the experimental setup, we next compare the PG-MMR method against the baselines on standard MDS datasets, evaluated by both automatic metrics and human assessors.

\vspace{0.05in}
\noindent\textbf{ROUGE~\cite{Lin:2004}.}
This automatic metric measures the overlap of unigrams (R-1), bigrams (R-2) and skip bigrams with a maximum distance of 4 words (R-SU4) between the system summary and a set of reference summaries.
ROUGE scores of various systems are presented in Table~\ref{tab:results_duc04} and~\ref{tab:results_tac11} respectively for the DUC-04 and TAC-11 datasets.

% \vspace{0.05in}
We explore variants of the PG-MMR method.
They differ in how the importances of source sentences are estimated and how the sentence importance affects word attention weights.
``\textbf{\emph{w/ Cosine}}'' computes the sentence importance as the cosine similarity score between the sentence and document vectors, both represented as sparse TF-IDF vectors under the vector space model. 
``\textbf{\emph{w/ SummRec}}'' estimates the sentence importance as the predicted R-L recall score between the sentence and the summary. 
A support vector regression model is trained on sentences from the CNN/Daily Mail datasets ($\approx$33K) and applied to DUC/TAC sentences at test time (see \S\ref{sec:our_approach}).
``\textbf{\emph{w/ BestSummRec}}'' obtains the best estimate of sentence importance by calculating the R-L recall score between the sentence and reference summaries. 
It serves as an upper bound for the performance of ``w/ SummRec.''
For all variants, the sentence importance scores are normalized to the range of [0,1].
``\textbf{\emph{w/ SentAttn}}'' adjusts the attention weights using Eq.~(\ref{equ:alpha_t_i_new_alternative}), so that words in important sentences are more likely to be used to generate the summary.
The weights are otherwise computed using Eq.~(\ref{equ:alpha_t_i_new}). 

As seen in Table~\ref{tab:results_duc04} and~\ref{tab:results_tac11}, our PG-MMR method surpasses all unsupervised extractive baselines, including SumBasic, KLSumm, and LexRank.
On the DUC-04 dataset, ICSISumm and DPP show good performance, 
but these systems are trained directly on MDS datasets, which are not utilized by the PG-MMR method. 
PG-MMR exhibits superior performance compared to existing abstractive systems.
It outperforms Opinosis and PG-Original by a large margin in terms of R-2 F-scores (5.03/6.03/\textbf{8.73} for DUC-04 and 5.12/6.40/\textbf{10.92} for TAC-11).
In particular, \textbf{\emph{PG-Original}} is the original pointer-generator networks with multi-document inputs at test time.
Compared to it, PG-MMR is more effective at identifying summary-worthy content from the input.
``w/ Cosine'' is used as the default PG-MMR and it shows better results than ``w/ SummRec.''
It suggests that the sentence and document representations obtained from the encoder-decoder model (trained on CNN/DM) are suboptimal, possibly due to a vocabulary mismatch, where certain words in the DUC/TAC datasets do not appear in CNN/DM and their embeddings are thus not learned during training.
Finally, we observe that ``w/ BestSummRec'' yields the highest performance on both datasets.
This finding suggests that there is a great potential for improvements of the PG-MMR method as its ``extractive'' and ``abstractive'' components can be separately optimized.

\begin{figure}[t]
\centering
\includegraphics[width=3in]{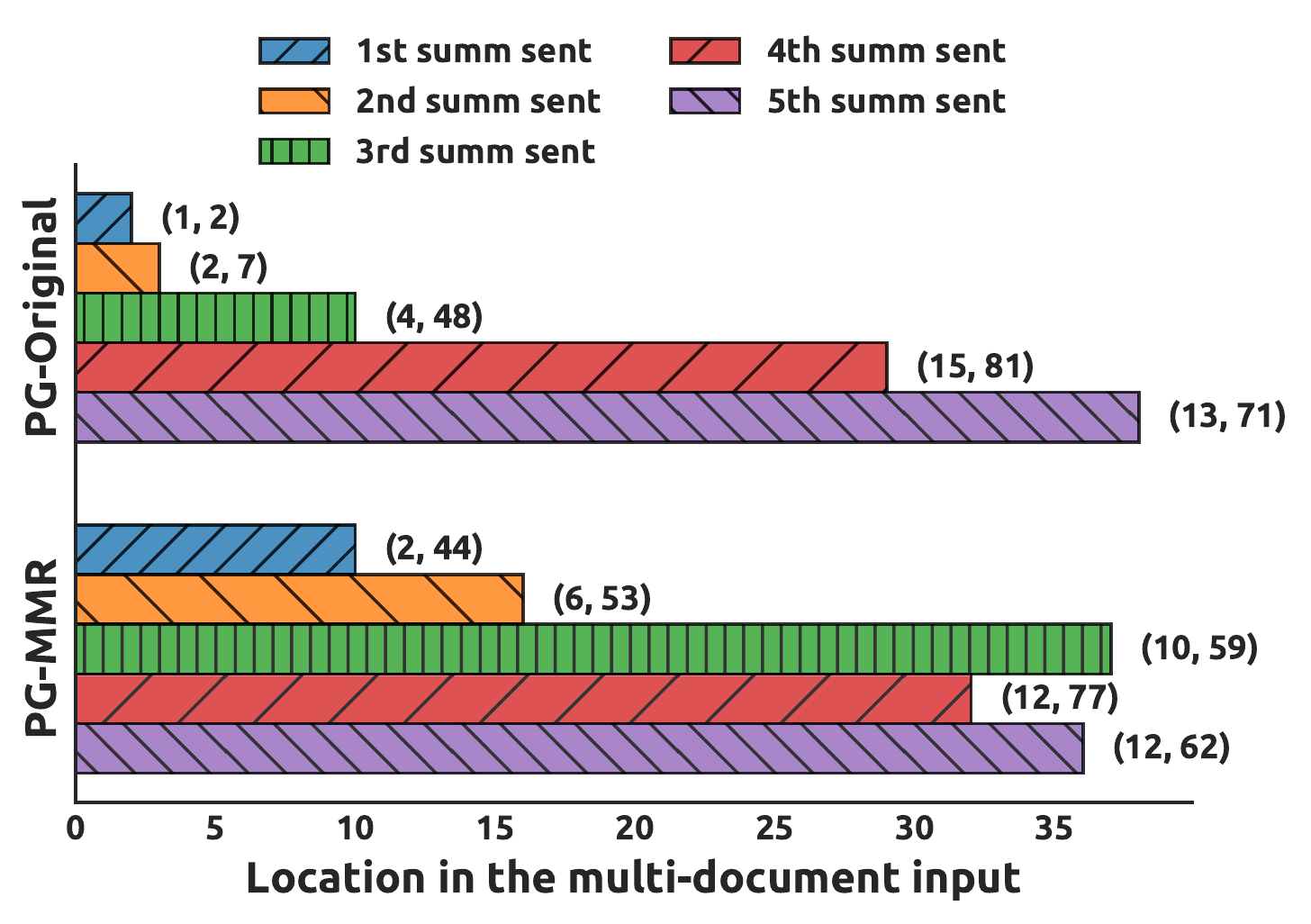}
\caption{The median location of summary n-grams in the multi-document input (and the lower/higher quartiles).
The n-grams come from the 1st/2nd/3rd/4th/5th summary sentence and the location is the source sentence index. (TAC-11)}
\label{fig:plot_pos}
\vspace{-0.1in}
\end{figure}

\begin{table}[t]
\setlength{\tabcolsep}{5pt}
\renewcommand{\arraystretch}{1.1}
\centering
\begin{small}
\begin{tabular}{|l|cccc|}
\hline
\textbf{System} & \textbf{1-grams}  & \textbf{2-grams}  & \textbf{3-grams}  & \textbf{Sent}\\
\hline
\hline
Extr+Rewrite & 89.37 & 54.34 & 25.10 & 6.65\\
PG-Original & 99.64 & 96.28 & 88.83 & 47.67\\
PG-MMR & 99.74 & 97.64 & 91.57 & 59.13\\
Human Abst.& 84.32 & 45.22 & 18.70 & 0.23\\
\hline
\end{tabular}
\end{small}
\caption{Percentages of summary n-grams (or the entire sentences) appear in the multi-document input. (TAC-11)
}
\label{tab:results_ngrams}
\vspace{-0.15in}
\end{table}

\begin{table*}[t]
\setlength{\tabcolsep}{6pt}
\renewcommand{\arraystretch}{1.1}
\centering
\begin{small}
\begin{tabular}{|l|ccc|rrrr|}
\hline
& \multicolumn{3}{c|}{\textbf{Linguistic Quality}} & \multicolumn{4}{c|}{\textbf{Rankings (\%)}}\\
\textbf{System} & \textbf{Fluency} & \textbf{Inform.} & \textbf{NonRed.} & \multicolumn{1}{c}{\quad\textbf{1st}} & \multicolumn{1}{c}{\quad\textbf{2nd}} & \multicolumn{1}{c}{\quad\textbf{3rd}} & \multicolumn{1}{c|}{\quad\textbf{4th}}\\
\hline
\hline
Extract+Rewrite & 2.03 & 2.19 & 1.88 & 5.6 & 11.6 & 11.6 & \textbf{71.2}\\
LexRank & \textbf{3.29} & 3.36 & 3.30 & 30.0 & 28.8 & 32.0 & 9.2\\
PG-Original & 3.20 & 3.30 & 3.19 & 29.6 & 26.8 & \textbf{32.8} & 10.8\\
PG-MMR & 3.24 & \textbf{3.52} & \textbf{3.42} & \textbf{34.8} & \textbf{32.8} & 23.6 & 8.8\\
\hline
\end{tabular}
\end{small}
\caption{Linguistic quality and rankings of system summaries. (DUC-04) 
}
\label{tab:results_mturk}
\vspace{-0.1in}
\end{table*}

\vspace{0.05in}
\noindent\textbf{Location of summary content.}
We are interested in understanding why PG-MMR outperforms PG-Original at identifying summary content from the multi-document input. 
We ask the question: where, in the source documents, does each system tend to look when generating their summaries?
Our findings indicate that PG-Original gravitates towards early source sentences, while PG-MMR searches beyond the first few sentences.

In Figure~\ref{fig:plot_pos} we show the median location of the first occurrences of summary n-grams, where the n-grams can come from the 1st to 5th summary sentence.
For PG-Original summaries, n-grams of the 1st summary sentence frequently come from the 1st and 2nd source sentences, corresponding to the lower/higher quartiles of source sentence indices.
Similarly, n-grams of the 2nd summary sentence come from the 2nd to 7th source sentences.
For PG-MMR summaries, the patterns are different. 
The n-grams of the 1st and 2nd summary sentences come from source sentences of the range (2, 44) and (6, 53), respectively.
Our findings suggest that PG-Original tends to treat the input as a single-document and identifies summary-worthy content from the beginning of the input, whereas PG-MMR can successfuly search a broader range of the input for summary content.
This capability is crucial for multi-document input where important content can come from any article in the set.

\begin{table*}[t]
\setlength{\tabcolsep}{5pt}
\renewcommand{\arraystretch}{1.1}
\begin{scriptsize}
\begin{fontppl}

\begin{minipage}[b]{0.5\hsize}\centering
\begin{tabular}[t]{|p{2.99in}|}
\hline
\textbf{Human Abstract}\\[1.1mm]
\textbullet\, Boeing 737-400 plane with 102 people on board crashed into a mountain in the West Sulawesi province of Indonesia, on Monday, January 01, 2007, killing at least 90 passengers, with 12 possible survivors. \\[1.1mm]                        
\textbullet\, The plane was Adam Air flight KI-574, departing at 12:59 pm from Surabaya on Java bound for Manado in northeast Sulawesi. \\[1.1mm]

\textbullet\, The plane crashed in a mountainous region in Polewali, west Sulawesi province. \\[1.1mm] 

\textbullet\, There were three Americans on board, it is not know if they survived. \\[1.1mm] 

\textbullet\, The cause of the crash is not known at this time but it is possible bad weather was a factor. \\[1.1mm] 
\hline
\hline
\textbf{Extract+Rewrite Summary}\\[0.4mm]

\textbullet\, Plane with 102 people on board crashes.\\[0.4mm]

\textbullet\, Three Americans among 102 on board plane in Indonesia.\\[0.4mm]

\textbullet\, Rescue team arrives in Indonesia after plane crash.\\[0.4mm]

\textbullet\, Plane with 102 crashes in West Sulawesi, killing at least 90.\\[0.4mm]

\textbullet\, No word on the fate of Boeing 737-400.\\[0.4mm]

\textbullet\, Plane carrying 96 passengers loses contact with Makassar.\\[0.4mm]

\textbullet\, Plane crashes in Indonesia , killing at least 90.\\[0.4mm]

\textbullet\, Indonesian navy sends two planes to carry bodies of five.\\[0.4mm]

\textbullet\, Indonesian plane carrying 102 missing.\\[0.4mm]

\textbullet\, Indonesian lawmaker criticises slow deployment of plane.\\[0.4mm]

\textbullet\, Hundreds of kilometers plane crash.\\[0.4mm]

\hline
\end{tabular}
\end{minipage}
\hfill
\begin{minipage}[b]{0.5\hsize}\centering
\begin{tabular}[t]{|p{2.99in}|}
\hline
\textbf{PG-Original Summary}\\[2mm]
\textbullet\, Adam Air Boeing 737-400 crashed Monday after vanishing off air traffic control radar screens between the Indonesian islands of Java and Sulawesi.\\[1.8mm]

\textbullet\, Up to 12 people were thought to have survived, with rescue teams racing to the crash site near Polewali in West Sulawesi , some 180 kilometres north of the South Sulawesi provincial capital Makassar.\\[1.8mm]

\textbullet\, It was the worst air disaster since Sept. 5, 2005, when a Mandala Airline's Boeing 737-200 crashed shortly after taking off from the North Sumatra's airport, killing 103 people.\\[1.8mm]

\textbullet\, Earlier on Friday, a ferry carrying 628 people sank off the Java coast.\\[2mm]
\hline
\hline
\textbf{PG-MMR Summary}\\[2mm]
\textbullet\, The Adam Air Boeing 737-400 crashed Monday afternoon, but search and rescue teams only discovered the wreckage early Tuesday. \\[1.9mm]

\textbullet\, The Indonesian rescue team arrived at the mountainous area in West Sulawesi province where a passenger plane with 102 people onboard crashed into a mountain in Polewali, West Sulawesi province. \\[1.9mm]

\textbullet\, Air force rear commander Eddy Suyanto told-Shinta radio station that the plane -- operated by local carrier Adam Air -- had crashed in a mountainous region in Polewali province on Monday. \\[1.9mm]

\textbullet\, There was no word on the fate of the remaining 12 people on board the boeing 737-400. \\[1.9mm]
\hline
\end{tabular}
\end{minipage}

\end{fontppl}
\end{scriptsize}
\caption{Example system summaries and human-written abstract. The sentences are manually de-tokenized for readability.}
\label{tab:results_output}
\vspace{-0.15in}
\end{table*}
% TAC 1105, test_004.txt

\vspace{0.05in}
\noindent\textbf{Degree of extractiveness.}
Table~\ref{tab:results_ngrams} shows the percentages of summary n-grams (or entire sentences) appearing in the multi-document input.
PG-Original and PG-MMR summaries both show a high degree of extractiveness, and similar findings have been revealed by See et al.~\shortcite{See:2017}.
Because PG-MMR relies on a handful of representative source sentences and mutes the rest, it appears to be marginally more extractive than PG-Original.
Both systems encourage generating summary sentences by stitching together source sentences, as about 52\% and 41\% of the summary sentences do not appear in the source, but about 90\% the n-grams do.
The Extract+Rewrite summaries (\S\ref{sec:experiments}), generated by rewriting selected source sentences to title-like summary sentences, exhibits a high degree of abstraction, close to that of human abstracts.

\vspace{0.05in}
\noindent\textbf{Linguistic quality.} To assess the linguistic quality of various system summaries, we employ Amazon Mechanical Turk human evaluators to judge the summary quality, including PG-MMR, LexRank, PG-Original, and Extract+Rewrite.
A turker is asked to rate each system summary on a scale of 1 (worst) to 5 (best) based on three evaluation criteria: \textbf{\emph{informativeness}} (to what extent is the meaning expressed in the ground-truth text preserved in the summary?), \textbf{\emph{fluency}} (is the summary grammatical and well-formed?), and \textbf{\emph{non-redundancy}} (does the summary successfully avoid repeating information?). 
Human summaries are used as the ground-truth. 
The turkers are also asked to provide an overall ranking for the four system summaries. 
Results are presented in Table~\ref{tab:results_mturk}.
We observe that the LexRank summaries are highest-rated on fluency.
This is because LexRank is an extractive approach, where summary sentences are directly taken from the input.
PG-MMR is rated as the best on both informativeness and non-redundancy.
Regarding overall system rankings, PG-MMR summaries are frequently ranked as the 1st- and 2nd-best summaries, outperforming the others.

\vspace{0.05in}
\noindent\textbf{Example summaries.}
In Table~\ref{tab:results_output} we present example summaries generated by various systems. 
PG-Original cannot effectively identify important content from the multi-document input. 
Extract+Rewrite tends to generate short, title-like sentences that are less informative and carry substantial redundancy.
This is because the system is trained on the Gigaword dataset~\cite{Rush:2015} where the target summary length is 7 words. 
PG-MMR generates summaries that effectively condense the important source content.

\section{Conclusion}
\label{sec:conclusion}

We describe a novel adaptation method to generate abstractive summaries from multi-document inputs. 
Our method combines an extractive summarization algorithm (MMR) for sentence extraction and a recent abstractive model (PG) for fusing source sentences. 
The PG-MMR system demonstrates competitive results, outperforming strong extractive and abstractive baselines.

% \section*{Acknowledgments}

% The acknowledgments should go immediately before the references.  Do
% not number the acknowledgments section. Do not include this section
% when submitting your paper for review. \\

\bibliography{new_ref,summ,abs_summ,fei}
\bibliographystyle{acl_natbib_nourl}

\end{document}